\documentclass[
]{ceurart}

\usepackage{listings}
\usepackage{longtable}
\begin{document}

\copyrightyear{2026}
\copyrightclause{Copyright for this paper by its authors.
  Use permitted under Creative Commons License Attribution 4.0
  International (CC BY 4.0).}

\conference{ETHICAIA'26: Workshop on Ethics in Clinical AI Applications at IJCAI-ECAI 2026, August 15--21, 2026, Bremen, Germany}

\title{Computational Phenomenology of Borderline Personality Disorder: A Comparative Evaluation of LLM-Simulated Expert Personas and Human Clinical Experts}

\author[1,2,3]{Marcin Moskalewicz}[%
email=marcin.moskalewicz@ideas.edu.pl
]
\cormark[1]
\address[1]{IDEAS Research Institute, Warsaw, Poland}
\address[2]{Poznan University of Medical Sciences, Poznan, Poland}
\address[3]{Maria Curie-Skłodowska University, Lublin, Poland}

\author[1,2]{Anna Sterna}[%
email=an.sterna@gmail.com
]

\author[1]{Karolina Drożdż}[%
email=karolina.drozdz@ideas.edu.pl
]

\author[1,4,5]{Kacper Dudzic}[%
email=kacper.dudzic@ideas.edu.pl
]
\address[4]{Adam Mickiewicz University, Poznań, Poland}
\address[5]{AMU Center for Artificial Intelligence, Poznan, Poland}

\author[6]{Marek Pokropski}[%
email=mpokropski@uw.edu.pl
]
\address[6]{University of Warsaw, Warsaw, Poland}

\author[7]{Paula Flores}[%
email=flores.psychoterapia@gmail.com
]
\address[7]{Zmiany.Psychoterapia, Warsaw, Poland}

\cortext[1]{Corresponding author.}

\begin{abstract}
  Building on a human-led thematic analysis of clinical life-story interviews ($>$ 150,000 words) with inpatients with Borderline Personality Disorder, this study examines the capacity of large language models (OpenAI's GPT, Google's Gemini, and Anthropic's Claude) to support qualitative clinical analysis. The models' interpretative potential was evaluated using a mixed-methods approach. Study A involved blinded and non-blinded judges in phenomenology and clinical psychology. The experts assessed the validity of AI-generated content using semantic congruence, Jaccard coefficients, and multidimensional validity ratings, including credibility, coherence, the substantiveness of results, and grounding in qualitative data. In Study B, neural methods were used to embed human- and model-generated theme descriptions in a multidimensional vector space. This approach provided an objectified computational measure of the difference between human and model semantics and linguistic style. In Study C, complementary non-expert evaluations were conducted ($N=115$) to examine the influence of thematic verbosity on the perception of human authorship and content validity. Overall, the results of AI analysis showed a highly variable overlap (0--58\%) with the human interpretation, while all models identified themes originally omitted by human researchers, proving their capacity to mitigate human bias. At the thematic level, external evaluators were unable to reliably distinguish human-authored themes from those generated by AI. In terms of content validity assessed against raw data by high-level experts in the blinded mode, the performance of Gemini 2.5 Pro was indistinguishable from that of humans. The comparison of semantic vector embeddings showed that its style was also the closest to humans.
\end{abstract}


\maketitle

\section{Introduction}

This project contributes to the emerging field of AI-enhanced qualitative phenomenology by testing the ability of large language models (LLMs) to reproduce human analysis of qualitative clinical data representing lived experience, including its nuances as well as overly abstract insights. Previous research demonstrates that LLMs can perform inductive thematic grouping of qualitative healthcare interviews \cite{Mathis2024}. Additionally, the analysis of human narratives with LLMs has recently been shown to be effective for predicting the level of psychopathology \cite{Ringwald2025}. AI-analyzed life narratives were also used to predict personality traits \cite{Oltmanns2025}. LLMs have proven useful at various stages of qualitative data analysis, including generating codes, clusters, and themes, and increasing the analysis's general trustworthiness \cite{Nguyen2025}.

One clear advantage, if treated as a subsidiary rather than the primary interpretative agent, is that LLMs may perform alternative thematic grouping and enhance research quality by addressing human limitations (such as fatigue and difficulties in synthesizing due to data overload) \cite{Valenzuela2019}. Human-led thematic analysis may thus be challenged and critically compared with the AI-enhanced one to reduce potential interpretative bias and increase the reliability of the outcomes. While human-led analysis may be prone to data overload and therefore limited in its ability to synthesize material effectively, LLMs may be better at pattern-seeking. Nevertheless, LLM outcomes may be restricted by the overreliance on face-value meanings. Therefore, the present study evaluates AI performance against a reference qualitative analysis grounded explicitly in phenomenological methodology, while accounting for raw data. 

The reference qualitative study \cite{sterna2025sense} explored self-disturbance in Borderline Personality Disorder (BPD). It aimed to empirically extract BPD-distinctive features of narrative identity and temporality, broadly conceptualized in phenomenological literature \cite{Fuchs2007,Schmidt2021}. The challenge of AI replication is that the reference study involved several strategies, potentially difficult to reproduce with LLMs, such as (i) the phenomenological principle of \textit{epoche}, or the suspension of judgment, as applied in qualitative research, which consists in ongoing self-reflection aimed at reducing hidden biases; (ii) focus on latent meanings as opposed to synthesis of superficial linguistic meanings; (iii) imaginative operations attempting to reconstruct the participants' ``lifeworld'' from the text  \cite{sterna2025sense}. The aim of the current research, comprising three studies (A, B, and C), was to validate the accuracy of AI analysis of the same data in a mixed-method design involving both blinded and non-blinded expert judges in phenomenology and clinical psychology, and laymen's public assessment\footnote{The original qualitative study research protocol was approved by both the hospital in which the research was held and the Poznan University of Medical Sciences Bioethical Committee (decision no. KB-367/23). The experimental part did not require external approval because all participants were co-authors of the study. Prolific participants provided informed consent acknowledging content warnings. Additional approval was not required as the study involved no deception}. 

\section{Study A: Methods}

\subsection{AI Qualitative Analysis}
The same set of instructions was sequentially applied to three (as of August 2025) state-of-the-art LLMs: OpenAI's GPT-4o, Google's Gemini 2.5 Pro, and Anthropic's Claude 4 Opus, all with default hyperparameter settings (for a detailed prompting design protocol---see Supplementary Materials). 

The models were first prompted to mimic the specific interpretative style of the primary human investigator of the original study (step 1). This was done using an advanced persona-prompting strategy guided by the actual, predefined professional characteristics of Anna Sterna (background, experience, courses in qualitative research, interpretative biases). This strategy is particularly relevant for tasks that require high openness, such as qualitative analysis \cite{Olea2025}. 

Further prompting followed the CO-STAR model \cite{Teo2023} comprising: context, objective, style, tone, audience, and response. As part of the context provision, models were familiarized with the published meta-synthesis of research on temporal experience in BPD \cite{Sterna2025Temporality} written by the authors of the original qualitative study (step 2). The aim was to reproduce the bias of human interpreters stemming from their background knowledge. Afterwards, the models were familiarized with a concrete textbook instruction on the thematic analysis of qualitative data used in the original study (step 3). Next, the models were familiarized with an original dataset in Polish \cite{sterna2025sense}, comprising Life Story Interviews \cite{McAdams1988} collected from 24, inpatient BPD individuals. Interviews lasted 50-70 minutes and were transcribed verbatim, resulting in a dataset of approximately 150,000 words (see Appendix A: Participants and Ethical Considerations). The models were given the same demographic background information on the patients as the researchers, along with the original research questions. The models were asked to perform thematic analysis in a style aligning with the previously sketched persona and with the responses intended for a particular scientific audience, namely, working in the field of phenomenological psychopathology (step 4). Three independent iterative rounds of step 4 were conducted for each model to ensure internal consistency 

Finally, the models were prompted to integrate the best elements of all three analyses and develop a refined thematic synthesis that addresses any flaws and better honors the phenomenological-experiential data (steps 5 and 6). The refined analysis, based on iterations' synthesis for each model, organized the results into final themes that constituted the dataset for the next stage of research (for the list of steps and verbatim prompt instructions, see Appendix C: Prompting Design Protocol for Study A). 

\subsection{Non-blinded Internal Expert Evaluation of Congruence}
Two experts (co-authors of the reference study) independently evaluated the AI outputs for congruence with their previous analysis regarding two criteria. 1) \textit{Semantic overlap}: absolute semantic overlap at the cross-thematic and sub-thematic level, comparing the contents of each human-generated theme against AI-generated ones (if all constituents of a human-generated theme were split into several themes generated by a particular AI model, such a theme was assessed as categorically congruent with the AI). The assessment had very high initial agreement ($>0.8$), and minor coding differences were consensually resolved. The final score was expressed as a percentage of overlap across all human-generated themes. 2) \textit{Thematic overlap}: main themes's overlap objectified with the Jaccard coefficient, where an AI-generated theme was judged as congruent and calculated into the interaction units when the semantic overlap was consensually assessed as covering $> 50\%$ of the original meaning. This coding tactics was employed to provide a categorical value since the Jaccard coefficient is a measure of similarity and diversity of two sets of data. It is defined as the intersection size divided by the union size, with scores ranging from 0 to 1, where 1 indicates more similarity. Since the limitation of both techniques is that any novel insights given by the AI would be, in principle, treated as false positives, all incongruent themes subsequently underwent manual verification against the raw qualitative data and were interpreted categorically as either nonsensical artifacts or novel contributions.

Discrepancies in scoring were expected due to differences in the evaluation methods used. Several methods were used because while absolute semantic overlap assumed that meanings could split across AI-generated themes, the main themes' semantic overlap lacked sensitivity and missed possible intersections at the subthematic level. Hence, the value of absolute overlap could be higher than the Jaccard for a given model when original meanings were more scattered between particular AI themes, whereas Jaccard could be potentially higher when the predominant congruence concerned the main themes only (given they passed the intended arbitrary 50\% overlap threshold).

\subsection{Non-Blinded Internal Experts and Blinded External Experts Evaluation of Content Validity}

\subsubsection{Content Validity Criteria.} To comparatively assess the more nuanced aspects of AI qualitative replication of human analysis, we used a set of criteria developed by others for assessing human qualitative studies, synthesized into eight categories. These were: substantiveness of contribution \cite{Tracy2010}, groundness in the data \cite{Charmaz2006,Rocco2010}, conceptual coherence \cite{Spencer2003,Yardley2000}, richness and complexity \cite{Spencer2003,Yardley2000,Braun2025}, theoretical integration \cite{Charmaz2006,Yardley2000}, multivocality \cite{Spencer2003}, practical usefulness \cite{Spencer2003,Yardley2000}, and credibility \cite{Tracy2010,Spencer2003}. Each criterion was applied to each analysis, humans and models, as a whole (i.e., all themes treated en bloc), which was assessed against it on a five-point Likert scale (with five representing a strong agreement).

\subsubsection{Evaluation.} Two additional external experts were involved, whose professional profiles mirrored those of the original interpreters (hence, four experts in total). Crucially, they familiarized themselves with the original raw interview data (ca. 150k words). Unlike the internal experts evaluating congruence, these external experts were blinded to the source of the analyses and the AI component, and were informed only that their task was to assess the validity of four distinct methods of qualitative data analysis (hence the blinded mode). The instruction and criteria were the same for the non-blinded internal experts and blinded external experts (eight categories on a five-point Likert scale). In addition, the external experts were asked to provide qualitative justification for their scores.

Krippendorff's Alpha \cite{Krippendorff2019} was employed via the K-Alpha Calculator \cite{Marzi2024} to assess the inter-rater reliability of all four coders, as it is particularly suited to studies with multiple raters and different levels of measurement. The analysis provided a reliability coefficient for the coding scheme, indicating the extent of agreement among raters beyond chance. The Kruskal-Wallis test was used to compare the validity scores of the four groups. Elasticity ratio, defined as $\frac{\text{\% change in words}}{\text{\% change in score}}$, was used to quantify the relationship between the number of words produced by humans and models and the summary experts' scores, to determine the potential benefit of more extensive wording. 

\subsection{External Experts AI Recognition}

The two external experts were informed that some previously evaluated analyses were AI-generated. They were asked to order the four outputs from the most to the least AI-probable.

\section{Study A: Results}
\subsection{Congruence}
\subsubsection{AI Qualitative Data} The models' outputs varied in terms of the number of themes and words. GPT distinguished 5 themes with 220 words ($M=44$ per theme), Gemini 7 themes with 681 words ($M=97.3$ per theme), and Claude 11 themes with 412 words ($M=37.5$ per theme). The original human analysis comprised 12 themes and 2031 words ($M=169$ words per theme). 

\subsubsection{Internal Experts} \textit {Semantic overlap} was highest for Claude (58\%), whose outputs covered 7 of the 12 original themes, second for Gemini (42\%), covering 5 of the original themes, and nonexistent for GPT. \textit{Thematic overlap} followed the same pattern, with the highest Jaccard coefficient for Claude (0.28), second for Gemini (0.27), and third for GPT (0.21). These results should be interpreted at the verge of low and moderate. 

All models presented new themes, which were subsequently verified against the raw qualitative data. One such theme per model proved insightful, as judged by the authors of the reference study, despite its absence in their original interpretation. GPT detected fragile moments of coherence in one of its themes. Gemini brought up the topic of otherness and yearning for normalcy, and Claude helped identify significant aspects of the bodily manifestations of the distress. \hyperref[tab:thematic_overlap]{Table 1} shows the examples of semantic and thematic overlap as well as new insights. A summary of data generated by LLMs is given in the Supplementary Materials. 

\subsection{Content Validity}

\subsubsection{Internal and External Experts} The Krippendorff's Alpha reached a satisfactory level $\ge 0.80$ as suggested by Krippendorff \cite{Krippendorff2019} for Claude (0.81) and Gemini (0.84), but was low for GPT (0.69) and mostly for human analysis (0.55). This was due to a skewed distribution of the human analysis, with coders varying between high-end scores of 4 and 5, which still counted as disagreement. The near-agreement within $\pm 1$ point between coders was approximately 85.71\%. To mitigate bias without settling the discrepancies, we treated the scores as independent values and averaged the data from all coders.

Human analysis reached the highest mean score of 4.56 ($Me = 4$; $Min-Max = 3--5$), second was Gemini with 4.21 ($Me = 4$; $Min-Max = 3--5$), third Claude with 3.68 ($Me = 4$; $Min-Max = 2--5$) and fourth GPT with 2.65 ($Me = 3$; $Min-Max = 1--4$). Kruskal-Wallis test showed significant differences between groups ($H = 60.2645, p < .00001$) with a mean rank for human analysis 93.47, Gemini 79.33, Claude 59.53, and GPT 25.67. Post-hoc comparisons using the Dunn-Bonferroni test indicated that mean rank scoring for both human analysis and Gemini were significantly higher than for GPT ($p < .0001$ both) and Claude ($p < .0001$ both). There was no significant difference between the human and Gemini analysis.

\subsubsection{Impact of Word Count.} Both the quantity of text produced (from 220 words of GPT to 2031 of humans) and the number of words per theme (from 44 in GPT to 169 in humans) were positively related to scores. The elasticity ratio with GPT as the baseline shows steady growth: Claude $0.445$; Gemini $0.281$; and human $0.088$. However, a cooling effect is visible, with the elasticity between Gemini and human analysis being $0.042$ only. 

\subsection{AI Recognition}

Ordering the four outputs from the most to the least AI-probable, the external experts noted that the probability is high for GPT and Claude but low for humans and Gemini. The ``humanness'' of Gemini aligned with its highest validity assessment score among the models (4.21 as given by all coders). It also aligned with the qualitative justification of the scores by external experts, to whom Gemini appeared (before they knew it was an LLM) as presenting a ``consistent narrative'', ``significant connections between various factors'', ``enhanced understanding of the qualitative data'', ``descriptions beyond surface-level'', and ``clear reference to existing theory and DSM classifications'', among others (see Supplementary Materials for more justifications). 

\begin{table}[!htbp]
\centering
\caption{Examples of overlaps between human qualitative study outputs and the models' analyses, including new insights by the models.}
\label{tab:thematic_overlap}
\definecolor{lightviolet}{RGB}{230, 220, 255} 
\definecolor{lightblue}{RGB}{220, 240, 255}
\definecolor{lightpink}{RGB}{255, 225, 230}
\footnotesize
\setlength{\tabcolsep}{4pt}
\renewcommand{\arraystretch}{0.98} 
\begin{tabular}{|p{0.1\textwidth}|p{0.11\textwidth}|>{\raggedright\arraybackslash}p{0.22\textwidth}|>{\raggedright\arraybackslash}p{0.47\textwidth}|}
\hline
\textbf{Model} & \textbf{Type} & \multicolumn{2}{c|}{\textbf{Theme Descriptions}} \\
\hline
\multirow{6}{*}{\textbf{GPT-4o}} 
 & \cellcolor{lightviolet}Semantic Overlap & \cellcolor{lightviolet}-- & \cellcolor{lightviolet}-- \\ \cline{2-4}
 & \cellcolor{lightblue}Thematic Overlap & \cellcolor{lightblue}\underline{Human:} \textbf{Disconnected episodic identifications} \newline discontinuity of heterogeneous self-experience & \cellcolor{lightblue}\underline{Model:} \textbf{Selfhood in Flux: Instability of Identity Over Time} \newline A recurring pattern was the unstable experience of self across time. Participants narrated an inability to recognize themselves in past or future projections, with shifting traits, values, and even memories. Selfhood emerged as something episodic, relationally entangled, and difficult to anchor across lived time. \\ \cline{2-4}
 & \cellcolor{lightpink}New Insights & \multicolumn{2}{>{\raggedright\arraybackslash}p{\dimexpr0.69\textwidth+2\tabcolsep+\arrayrulewidth\relax}|}{\cellcolor{lightpink}\textbf{Fragile Moments of Coherence and Future Orientation} \newline Amid fragmentation, some participants described glimpses of narrative clarity or hope. Often tied to treatment, learning, or relationships, these moments reflected efforts to reclaim ownership over time and self even if temporarily fragile or aspirational in nature.} \\ \hline

\multirow{9}{*}{\textbf{Gemini 2.5}} 
 & \cellcolor{lightviolet}Semantic Overlap & \cellcolor{lightviolet}\underline{Human:} \textbf{Harmful historicity} \newline identification of harmful aspects of one's life story & \cellcolor{lightviolet}\underline{Model:} \textbf{The Enduring Shadow of an Adverse Early Environment} \newline This theme highlights the profound and pervasively felt impact of early life experiences, which were frequently described as characterized by various forms of trauma, neglect, sustained parental conflict, or a fundamental lack of emotional safety and consistent nurturance. For these individuals, the past is not a static backdrop but an active, intrusive, and often overwhelmingly negative force. \\ \cline{2-4}
 & \cellcolor{lightblue}Thematic Overlap & \cellcolor{lightblue}\underline{Human:} \textbf{Blurred self-concept} \newline difficulties synthesizing felt changeability & \cellcolor{lightblue}\underline{Model:} \textbf{The Unstable and Fragmented Experience of Self} \newline This theme speaks to a core identity disturbance characterized by a painful and confusing lack of a stable, cohesive sense of self. Participants frequently voiced an inability to answer ``who am I?'' describing an internal experience of emptiness, of being ``nikim'' (nobody), or of being like a chameleon, compelled to adopt different personas to fit external contexts without a felt sense of an authentic, enduring core. \\ \cline{2-4}
 & \cellcolor{lightpink}New Insights & \multicolumn{2}{>{\raggedright\arraybackslash}p{\dimexpr0.69\textwidth+2\tabcolsep+\arrayrulewidth\relax}|}{\cellcolor{lightpink}\textbf{The Persistent Sense of Otherness and a Yearning for Normalcy} \newline This theme highlights a pervasive and painful felt sense of being fundamentally different, alien, flawed, or an outcast in relation to others and to the perceived social world. This experience of ``otherness'' is often deeply ingrained, contributing to feelings of isolation, shame, and inadequacy. It coexists with, and is often intensified by, a powerful and poignant yearning for a life that feels ``normal''---a life characterized by stability, genuine connection, acceptance, a sense of purpose, and inner peace. This desired ``normalcy'' frequently appears as an idealized and tragically unattainable state.} \\ \hline

\multirow{6}{*}{\textbf{Claude}} 
 & \cellcolor{lightviolet}Semantic Overlap & \cellcolor{lightviolet}\underline{Human:} \textbf{Lack of belonging} \newline difficulties identifying with social groups & \cellcolor{lightviolet}\underline{Model:} \textbf{Existential Isolation Despite Desperate Connection-Seeking} \newline A profound sense of not belonging permeates participants' accounts---feeling alien in the world while simultaneously desperately seeking connection. This creates a painful paradox where isolation feels safer yet unbearable. \\ \cline{2-4}
 & \cellcolor{lightblue}Thematic Overlap & \cellcolor{lightblue}\underline{Human:} \textbf{Hindered self-development} \newline sense of being trapped in non-progressing past & \cellcolor{lightblue}\underline{Model:} \textbf{Repetition Compulsion and Cyclical Existence} \newline Life is experienced as an endless repetition of the same painful patterns, relationships, and failures. Participants describe feeling trapped in cycles they can recognize but feel powerless to break, contributing to a sense of temporal stasis. \\ \cline{2-4}
 & \cellcolor{lightpink}New Insights & \multicolumn{2}{>{\raggedright\arraybackslash}p{\dimexpr0.69\textwidth+2\tabcolsep+\arrayrulewidth\relax}|}{\cellcolor{lightpink}\textbf{Affective Extremes and Embodied Emotional Overwhelm} \newline Emotional experience is characterized by intense, immediate, and somatic manifestations that feel uncontrollable and all-consuming. Participants describe living between extremes with no stable middle ground---rapid shifts between euphoria and despair, idealization and devaluation, creating an exhausting inner landscape where emotions are experienced as physical forces that completely overtake consciousness.} \\ \hline
\end{tabular}
\end{table}

\section{Study B: Methods}
\subsection{Computational Semantic and Stylistic Match}
A follow-up experiment explored the similarity between human- and LLM-created theme summaries through converting them into semantic vector embeddings. It enabled an objectified computational measure of the degree of similarity between human and AI-generated qualitative analysis and its comparison with the human evaluators' assessments. Using the \texttt{Qwen3-Embedding-8B}\footnote{\url{https://huggingface.co/Qwen/Qwen3-Embedding-8B}} model \cite{zhang2025qwen3embeddingadvancingtext}, we created semantic vector embeddings for the summary of each theme proposed by each agent. Model choice was motivated by its all-around performance on the MMTEB benchmark \cite{enevoldsen2025mmtebmassivemultilingualtext} for English language texts, its open-source nature, and its provider not being related to any of the evaluated LLMs. 

We also created additional embeddings using the \texttt{StyleDistance}\footnote{\url{https://huggingface.co/StyleDistance/styledistance}} model \cite{patel-etal-2025-styledistance}. Unlike a standard text embedding model, \texttt{StyleDistance} creates embeddings that focus on capturing the particulars of the linguistic style of the inputs as opposed to their semantic content \cite{patel-etal-2025-styledistance}; this enables a comparison of differences between human- and LLM-created text from a different angle. We labeled the two contrasting approaches \textit{semantics-first} and \textit{style-first}.

Given the embeddings, we first calculated within-group and between-group theme-level mean cosine similarities. Within-group similarity is the mean cosine over all unordered theme paragraph pairs within a group (excluding self-pairs), whereas between-group similarity is the mean cosine over all theme paragraph pairs across the two groups (e.g., 77 pairs for 7 Gemini pairs and 11 Claude pairs). In parallel, we also conducted $L^2$ normalization on the embeddings, followed by dimensionality reduction to a two-dimensional projection with UMAP \cite{mcinnes2020umapuniformmanifoldapproximation} using \texttt{scikit-learn}\footnote{\url{https://github.com/scikit-learn/scikit-learn}} \cite{JMLR:v12:pedregosa11a} (hyperparameters: \texttt{n\_components}=\texttt{2}, \texttt{n\_neighbors}=\texttt{15}, \texttt{min\_dist}=\texttt{0.1}, \texttt{metric}=\texttt{cosine}, \texttt{random\_state}=\texttt{42}, rest default). 

\section{Study B: Results}

\subsection{Computational Semantic and Stylistic Match}
\begin{figure*}[t]
\centering
\includegraphics[width=0.85\textwidth]{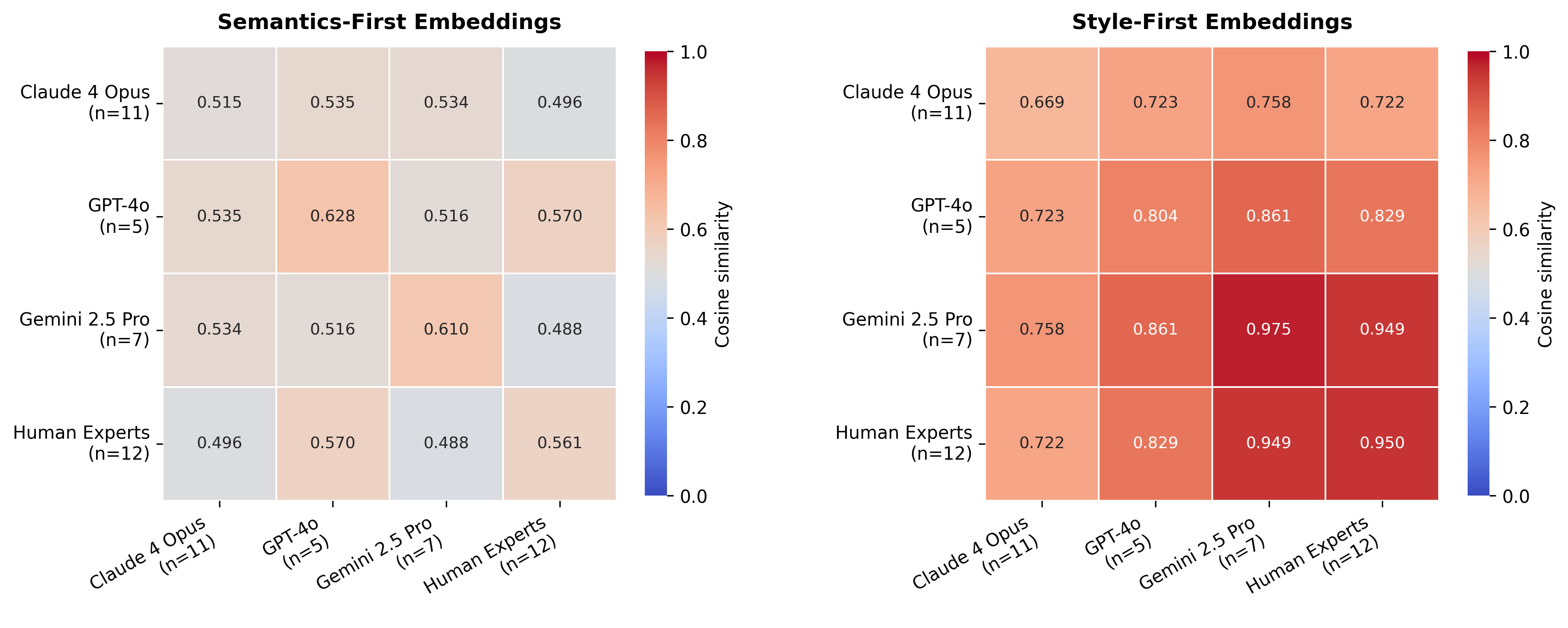}
\caption{Pairwise cosine similarities for the text embeddings of theme descriptions in two variants: semantics-first and style-first.}
\label{fig:cosine_heatmaps}
\end{figure*}

\begin{figure*}[t]
\centering
\includegraphics[width=0.85\textwidth]{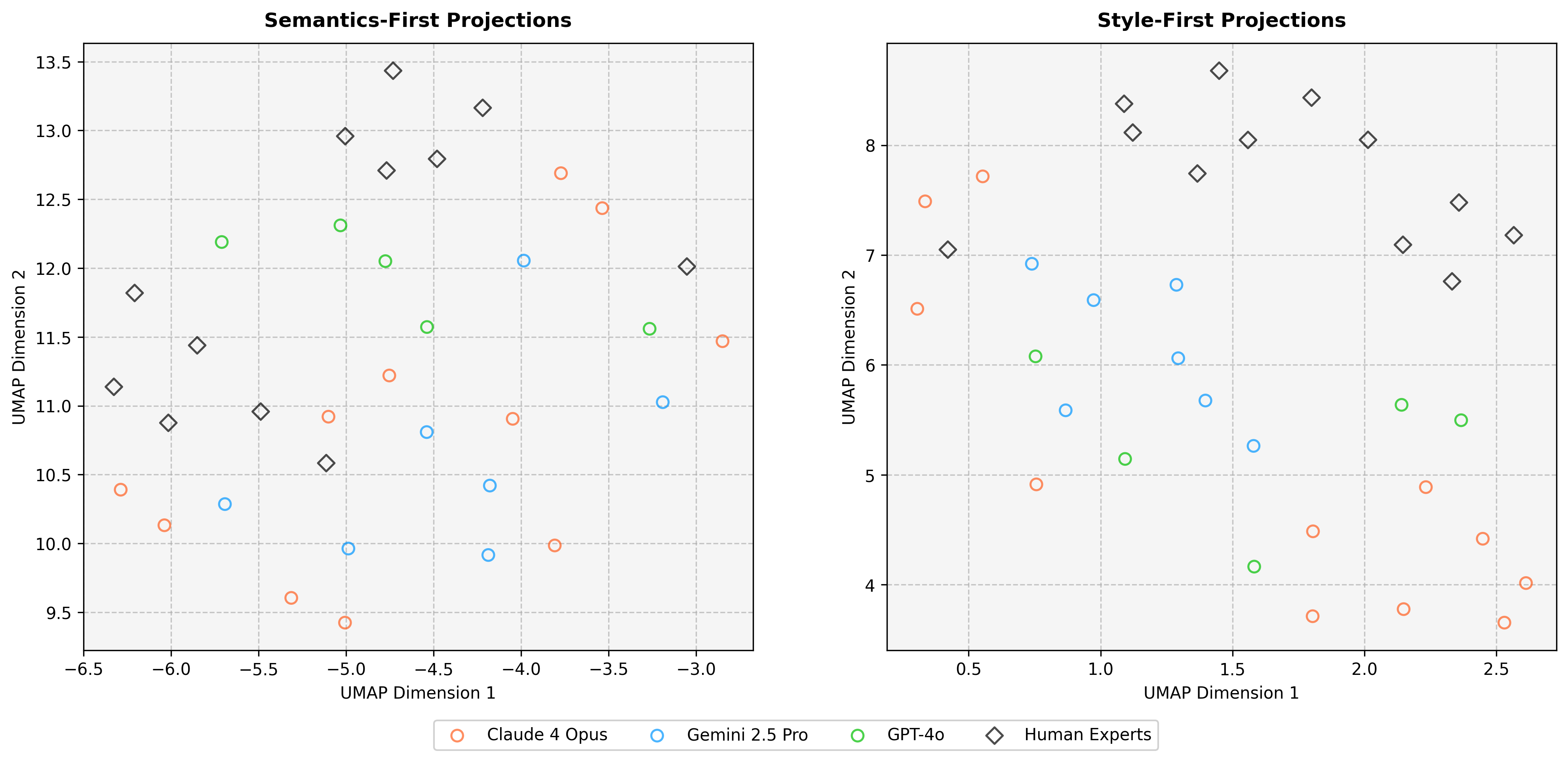}
\caption{UMAP projections of the text embeddings of theme descriptions in two variants: semantics-first and style-first.}
\label{fig:umap_comparison}
\end{figure*}

The theme-level mean cosine similarities are visualized in \hyperref[fig:cosine_heatmaps]{Figure \ref{fig:cosine_heatmaps}}, whereas the UMAP projections in \hyperref[fig:umap_comparison]{Figure \ref{fig:umap_comparison}}. For the semantics-first embeddings, both within-group and between-group similarities are situated in narrow bands; 0.515--0.628 and 0.488--0.570, respectively. The gap between the among-model mean of 0.529 $\pm$ 0.009 and the human-to-model mean of 0.518 $\pm$ 0.037 is negligible at $+$0.011. It can thus be said that semantically, the themes delineated by both models and humans are similar. The between-theme variety is, in contrast, relatively high, which accounts for the similarity score values remaining below 0.7; this can stem from the freedom given to agents in theme creation, as well as the complicated clinical cases the themes were ultimately based on.

The style-first embeddings paint a different picture. Within-group similarities vary from a relatively low 0.669 for Claude to Gemini's 0.975 and a close 0.950 for humans. Between-group similarities are situated within a slightly narrower range, starting from 0.722 for Claude-human similarity to 0.949 for Gemini-human. The among-model mean of 0.781 $\pm$ 0.059 and the human-to-model mean of 0.833 $\pm$ 0.093 are both markedly higher than in the semantics-first case, while the gap between them is also larger in magnitude at $-$0.0524. 

Overall, the style-first findings suggest that among the agents, Claude exhibits the biggest stylistic spread overall, illustrated by its within-group value of 0.669 (lowest overall) and between-group values all below 0.8. In contrast, Gemini can retain a consistent and human-like style very well, with the highest within-group value of 0.975 and a very high value of 0.949 similarity to humans. Finally, GPT is situated in the middle of those two extremes. Somewhat surprisingly, humans are closer to models in terms of style than models are to each other on average---the gap between the among-model and human-to-model means is mostly driven by Claude pulling the among-model score down.

The cosine-similarity means are corroborated by the UMAP projections. Inspecting the visualizations, semantics-first projections are rather uniformly dispersed, with a lot of overlap between agents, whereas style-first projections are characterized by a much stricter clustering together of embeddings created by a particular agent (except for Claude's, which exhibits a notably weaker clustering tendency). Additionally, higher-level style-first clusters of human- and LLM-created embeddings can be visually distinguished, although Gemini's style remains the closest to the human expert's.

\section{Study C: Methods}
\subsection{Validity of Themes}
\subsubsection{Public Evaluation} In the third study, a larger group of evaluators recruited via Prolific was involved ($N = 115$, 54.8\% male, 45.2\% female, mean age 35.09 years, $SD = 12.23$, ethnically diverse, with 40.9\% identifying as Black, 37.4\% as White, 7.8\% as Asian, and 7.8\% as Mixed ethnicity). All participants received monetary compensation for their participation. Eligibility was limited to self-reported fluency in English. Participants were presented with 10 themes from both the original qualitative study and the AI qualitative analysis in the current research, displayed on a single page. All themes were standardized by correcting grammatical errors and removing formatting artifacts typical of AI outputs. The stimuli were randomly drawn from a total pool of 35 themes created by human experts ($n = 12$), or generated by LLMs (Claude $n = 11$; Gemini $n = 7$; GPT $n = 5$). The survey was configured to present themes evenly, ensuring each theme was displayed to participants around 33 times. Participants evaluated each theme using a five-point Likert scale (with five representing a strong agreement) regarding semantic match (``The theme title matches the theme content'').

The key methodological difference between expert evaluation in Study A and public evaluation in Study C was, therefore, both the validity criteria and the fact that experts judged the contents' representational value \textit{en masse} and against the whole corpus of raw qualitative data, whereas public evaluators focused on randomly shuffled separate themes and judged the representational value of theme titles against theme descriptions.

\subsection{Perceived Authorship (Human or AI)}
\subsubsection{Public Evaluation.} The group of public evaluators was asked to assess the randomly assigned themes in terms of their perceived authorship using a five-point Likert scale (``The theme is generated by AI and not by a human expert''), where five represented strong agreement.

As with the previous task, the key methodological difference regarding Study A was that the experts had access to raw data, against which they could judge the representational value of both human- and AI-generated content in full, whereas public evaluators only assessed randomly assigned thematic descriptions, independently of their source.

\section{Study C: Results}
\subsection{Validity of Themes}
\subsubsection{Public Evaluation.} Descriptive statistics indicated differences in the perceived semantic match between theme titles and theme content across agents. Themes generated by Claude received the highest mean ratings for semantic match ($M=4.17, SD=1.06$), followed by Gemini ($M=4.06, SD=1.14$), human experts ($M=3.95, SD=1.10$), and GPT ($M=3.68, SD=1.25$). Due to the non-normal distribution of the data (Shapiro-Wilk $p < .001$ for all groups), a Kruskal-Wallis $H$ test was conducted. The analysis revealed a statistically significant difference in semantic match scores between the groups ($H(3) = 24.45, p < 0.001$). Post-hoc pairwise comparisons using Dunn's test (Bonferroni-adjusted) indicated that the semantic match for Claude's themes was rated significantly higher than for both human experts ($p < .01$) and GPT ($p < .001$). Similarly, Gemini's themes were perceived as having a significantly better semantic match than GPT's ($p < .01$) but did not differ significantly from those generated by human experts ($p = .49$). Notably, there was no significant difference between theme ratings of human experts and GPT ($p = .21$).

\subsubsection{Impact of Word Count.} Given that the themes varied significantly in terms of length, a Pearson correlation was conducted to examine whether word count influenced participant ratings. The analysis revealed no significant correlation between word count and perceived semantic match ($r=-.167, p=.339$).

\subsection{Perceived Authorship (Human or AI)}

\subsubsection{Public Evaluation.} Mean scores for all agents clustered around the midpoint of the scale, reflecting high uncertainty: Gemini ($M=3.02, SD=1.61$), Human ($M=2.79, SD=1.59$), GPT ($M=2.61, SD=1.49$), and Claude ($M=2.57, SD=1.59$). The Kruskal-Wallis test indicated a significant main effect ($H(3)=12.60, p < .01$) of the agent on perceived authorship ratings. However, post hoc analysis revealed that this significance was driven entirely by the perception of Gemini as significantly more ``AI-like'' than Claude ($p < .01$). Crucially, participants did not distinguish between human expert- and LLM-generated themes in terms of perceived authorship, with no statistically significant differences observed across conditions (all $p > .05$). 

A significant moderate positive correlation was found between word count and perceived authorship ($r=.344, p=.043$). This indicates that participants likely employed verbosity as a heuristic---longer themes were more likely to be perceived as AI-generated, regardless of their actual source. \hyperref[tab:results_summary]{Table \ref{tab:results_summary}} summarizes the results regarding semantic and thematic overlap, validity, and AI recognition from Study A and semantic match and perceived authorship from Study C.

\begin{table}[!htbp]
    \centering
    \caption{Summary of results; descriptive statistics, validity scores, and comparative metrics across human and LLM personas.}
    \label{tab:results_summary}
    \small
    \setlength{\tabcolsep}{3pt}
    \begin{tabular}{lccccc}
        \toprule
        \textbf{Metric experts} & \textbf{Human} & \textbf{GPT-4o} & \textbf{Gemini Pro 2.5} & \textbf{Claude 4 Opus} & \textbf{Word count ($r$)} \\
        \midrule
        Words / Themes / WPT & 2031 / 12 / 169 & 220 / 5 / 44 & 681 / 7 / 97.3 & 412 / 11 / 37.5 & -- \\
        Semantic / Thematic overlap & -- & 0 / .21 & 42\% / .27 & 58\% / .28 & -- \\
        \makecell[l]{Validity Score\\\hspace{1.5em} $M$($Me$)$Min-Max$} & 4.56(4)3--5$^{*\dagger}$ & 2.65(3)1--4$^{*}$ & 4.21(4)3--5$^{*\dagger}$ & 3.68(4)2--5$^{*}$ & --\\
        AI Recognition & Low & High & Low & High & -- \\

  \toprule
        \textbf{Metric public} & \textbf{Human} & \textbf{GPT-4o} & \textbf{Gemini Pro 2.5} & \textbf{Claude 4 Opus} & \textbf{Word count ($r$)} \\

        \midrule
        Validity of themes $M$($SD$)$^{\S}$ & 3.95 (1.10)$_{\text{b,c}}$ & 3.68 (1.25)$_{\text{c}}$ & 4.06 (1.14)$_{\text{a,b}}$ & 4.17 (1.06)$_{\text{a}}$ & $-.17$ \\
        \makecell[l]{Perceived Authorship\\\hspace{1.5em} $M$($SD$)$^{\parallel}$} & 2.79 (1.59)$_{\text{a,b}}$ & 2.61 (1.49)$_{\text{a,b}}$ & 3.02 (1.61)$_{\text{a}}$ & 2.57 (1.59)$_{\text{b}}$ & $.34^{\ddag}$ \\
        \bottomrule
    \end{tabular}
    
    \medskip
    \begin{minipage}{0.95\textwidth} 
    \footnotesize
    \textit{Note.} Metrics above the middle line denote Expert Evaluations; metrics below denote Public Evaluations. WPT = Words Per Theme. \\
    $_{\text{a,b,c}}$ Means within a row sharing the same subscript are not significantly different at the $p < .05$ level (Dunn's test with Bonferroni correction). Means with different subscripts are significantly different.\\
    $^{*}$ Validity Score: Kruskal-Wallis $H = 60.26$, $p < .00001$.\\
    $^{\dagger}$ Dunn-Bonferroni post-hoc: Human vs. GPT \& Claude ($p < .00001$); Gemini vs. GPT \& Claude ($p < .00001$).\\
    $^{\ddag}$ Word count ($r$): Pearson's $r$ significant at $p < .05$.\\
    $^{\S}$ Semantic Match: Kruskal-Wallis $H(3)=24.45$, $p < .001$.\\
    $^{\parallel}$ Perceived Authorship: Kruskal-Wallis $H(3)=12.60$, $p < .01$.
    \end{minipage}
\end{table}
\section{Discussion}

This study demonstrated the limited efficacy of three mainstream LLMs in conducting thematic analysis of BPD autobiographical narratives. While the results partly mirror the results of human consensual interpretation, the analytical accuracy remains limited.  

In the expert evaluation part of this research (Study A), Gemini performed best, with no statistically significant differences from human analysis. This was further reflected in its attribution as human-authored, which was due to its interpretive nuance and deep clinical insight as reported by external experts who were not aware that they were dealing with AI. The same experts considered GPT's worst performance to be due to theoretical overdetermination and a too-rigid focus on temporality, both of which led them to correctly recognize it as artificial. Meanwhile, evaluators in Study C, operating at the face-level value of theme descriptions, were unable to consistently and reliably distinguish human-authored from AI-generated themes. Moreover, some of the AI-generated content proved insightful to the original interpreters. These results are promising, suggesting a potential advantage of AI analysis in targeting omissions and thereby increasing the sensitivity of human qualitative insights. 

The results of the computational linguistic exploration in Study B, which focused on the semantic and stylistic match between human- and AI-conducted thematic analyses, provide additional context for the expert judgments. It can be asserted that the semantic distribution of themes is comparable across agents and that the expert persona-induced LLMs can align well semantically with the original human experts' analytic outputs. In other words, LLMs can mimic the interpretative content of mental health professionals at a face-value level (that is, regardless of their correspondence to raw data contents) based on an in-prompt description of the person alone. In contrast, the themes differ in terms of linguistic style, with a substantial variability in the ability of LLMs to mimic the style of human professionals. Expert preference for Gemini is more likely to be explained through the similarity of its generations to human answers in terms of linguistic style, as its embeddings were the closest match to human embeddings stylistically (although still being clearly a part of the LLM cluster), while being similarly dispersed semantically to other agents. It is possible that a convincingly human (compared to other models) presentation of the qualitative themes influenced the experts to consider their contents as more likely to be human-produced as well; however, this hypothesis warrants further investigation. 

Despite the high overall computational semantic similarity among all agents, GPT's worst expert-rated performance could still be attributable to singular mentions of specific topics or particular phrasings atypical for humans, both of which are not directly reflected in the (mathematically approximate) values of its embeddings. On the whole, the computational results point toward a performance gap in the ability of LLMs to act as stand-ins for human experts in the evaluated task: they can effectively support humans in creating human expertise-aligned contents of qualitative clinical analyses (the semantic content), but by themselves are less effective and more varied in their ability to reliably present the results in a convincingly human-like manner (the linguistic style). The finding concerning semantic convergence between LLM- and human-generated themes is corroborated by recent studies on clinical qualitative contexts \cite{Kivity2026,PLOSblinded2026,Grover2026}. The literature is also consistent in finding that LLM outputs are invariably stylistically distinguishable from human writing \cite{Reinhart2025}. Unlike prior work, which, e.g., approximated style through surface readability metrics \cite{Grover2026}, our persona-prompted design, combined with dedicated style embeddings, shows that the stylistic gap can be model-dependent, with one model closely approaching the style of the simulated human expert.

Despite Gemini's human-like performance in Study A, the stability of results was low, with significant variability in the outputs of different AI models regarding theme choice and phrasing. While variability is commonly observed in human qualitative studies, differences in variability require empirical verification through systematic comparisons of multiple AI- and human-generated research outputs. Another issue is the limited methodological transparency of autoregressive AI models. While human self-reflection also does not guarantee complete transparency, it enables the identification of personal biases (either individually or in collaboration with the research team). In contrast, the latent AI biases are more difficult to control and address. At the same time, all models successfully identified omitted themes, indicating the limits and hidden biases of human researchers working with large datasets, particularly in synthesizing all relevant issues. AI-generated analyses were likely effective in this respect also due to the lack of cognitive fatigue. 

The number of words, meanwhile, was important but not crucial to the quality of the content produced. The quantity of text was related to higher scores from external experts with access to raw data, but not when judged on the face value of the thematic content in the public assessment. In the latter case, longer themes were more likely to be perceived as AI-generated, regardless of their actual source. In the experts' assessment, the cooling effect in the elasticity score suggests that while more words indeed lead to a higher score, the extensive human analysis provides little benefit, also given the lack of a significant difference with the Gemini content validity score. The most efficient score gain occurs at word counts between the LLMs. It could be argued that the work required to add significantly more words, as in the original human analysis, far outweighs the tiny improvement in the score. 

All in all, the results indicate that persona-based, AI-enhanced qualitative analysis offers a promising path to increasing the reliability of human-led analysis, beyond the previously applied, e.g. codebook-level enhancements \cite{bryda2024words}. The analytic strategy we used proved to increase the granularity of analyses and to identify human biases. Thus, if carefully controlled, it provides a methodological upgrade that strengthens the rigor of qualitative analysis.

In the original reference study, the results of the thematic analysis were further refined in a phenomenological mode to identify invariant structures of experience. Whether the studied LLMs are capable of producing such a complex and integrated structure remains uncertain. We must bear in mind that, while the AI-generated thematic outputs appeared cohesive, this was largely due to the interpretive efforts of human evaluators, who judged them in terms of meaningful wholes. The LLMs themselves provided only a constellation of thematically related elements. 

\subsection{Limitations} 
The number of external experts (two) in study A was small by the standards of machine learning evaluations. This limiting factor was deliberate, however, since the raw verbatim transcripts were over 150,000 words. Analyzing such a vast amount of data against several qualitative interpretations was a significant cognitive toll that required dozens of hours from high-level experts. 

On the other hand, although the number of public evaluators in Study C was relatively large, they had no access to the raw data, a limitation imposed by both logistical constraints and the group's limited high-level competence. Unlike the clinical experts, these untrained participants were only assessing 'layman semantics' - focusing on general coherence rather than the deep phenomenological validity that requires specialized training. The intended trade-off was that while Study A provided depth, Study C provided breadth, but their results should be treated independently. 

Another limitation is that alternative (or dedicated) prompting strategies could severely affect the results. Differing prompting approaches between the models might also render their outputs more relevant. In the current study, the prompting had to be consistent across all models for comparison, leaving room for improvement in a real research setting.

It also has to be acknowledged that LLMs are prone to generating erroneous and biased content \cite{Huang2025}, including in clinical applications \cite{Omar2025review, Omar2025natmed}. As such, their bias-mitigating role described in this work is limited to a particular intervention targeting a specific, largely non-overlapping human bias type (omissions due to fatigue and data overload), ultimately conditional on verification against raw data. 

Finally, regarding LLM-specific data safety risks, closed-source models were prioritized in the study due to the ease and the technical and financial feasibility of use in real clinical contexts. Nevertheless, in optimal deployment scenarios, locally hosted open-source models are strongly recommended, as they eliminate the risk of transmitting sensitive medical data to external providers.

\section{Conclusions} 
This paper shows that AI analysis of clinical autobiographical interviews with vulnerable individuals with troubled pasts may be indistinguishable from the interpretation by human qualitative research experts, though with significant variability. The possible overlap is not merely superficially semantic at the thematic interpretative level, but also valid with respect to the reference of human and AI-generated themes to raw data when assessed in a blinded mode. While the thematic congruence between human and AI results ranges from low to medium, the models can spot themes missed by clinical experts, thereby increasing the sensitivity of qualitative insights. Thus, LLMs may be used as a tool to mitigate interpretative bias, given that they are subject to manual scrutiny of results by human interpreters. 

Consequently, constructive meta-prompting techniques, which do not require fine-tuning of models, make the use of AI for the interpretation of nuanced first-person narratives of borderline patients (and possibly other personality disorder patients) feasible in many respects. Nevertheless, the high variability in the quality of AI outputs remains the main problem, since, in order to play a supportive role in the analysis of first-person qualitative data, the outputs of commercial large language models still need to be carefully assessed against the raw data rather than taken at face value. 

%
%



\begin{acknowledgments}
The initial research was funded by the National Science Center, Poland [No. 2021/42/E/HS1/00106] and complemented by the Institute IDEAS, Warsaw.  
\end{acknowledgments}

\section*{Declaration on Generative AI}
The authors have not employed any Generative AI tools except for the experiments reported in the paper.

\bibliography{sample-ceur}


\newpage

\appendix

\section{Participants and Ethical Consideration for the Reference Qualitative Analysis}

\textbf{Participants.}
Participants (N = 24), aged 18-29 y.o., white, Polish, all hospitalized at the Neurosis and Personality Disorders Ward in Miedzyrzecz, Poland. They were diagnosed with borderline personality disorder according to ICD-10 criteria. Exclusion criteria comprised: suicidal crisis, psychotic decompensation or substance abuse, all diagnosed 3 months prior to interview. This was done in order to control an influence of significant psychiatric comorbidities on the data. Personality disorder symptoms severity was assessed using LPFS-BF 2.0., the final study group comprised four levels of PD severity: mild (N = 5), moderate (N = 6), severe (N = 9), extreme (N = 4). Participants of various severity levels were enrolled in order to identify patterns potentially cross-cutting variations. 
Participants were all recruited voluntarily. All the interviews were conducted in Polish by a clinical psychologist, being also a certified psychotherapist, skilled to address any, potential distress along the interview.
Detailed description of the studied group, methods and analytic procedures used can be found in the original, referential study \cite{sterna2025sense}.

\textbf{Ethical considerations.}
The original, qualitative research protocol obtained approval from the Bioethical Committee at Poznan University of Medical Sciences decision no: KB-367/23. All the participants signed an informed consent form and received detailed information regarding the study aims, interview procedures and data analysis. The dataset was carefully anonymized prior to analysis, securing participants’ privacy. The use of AI was limited exclusively to the analysis of previously anonymized transcripts. To minimize the risk of algorithmic bias or overinterpretation, AI outputs were continuously, critically reviewed by the research team and carefully compared with the original material.

\section{Human Experts' Reference Qualitative Analysis}

\renewcommand{\arraystretch}{1.3} %
\begin{longtable}{|p{0.30\textwidth}|p{0.65\textwidth}|}
\hline
\textbf{Theme} & \textbf{Description} \\
\hline
\endfirsthead

\multicolumn{2}{c}%
{\textit{Continued from previous page}} \\
\hline
\textbf{Theme} & \textbf{Description} \\
\hline
\endhead

\hline \multicolumn{2}{c}{\textit{Continued on next page}} \\
\endfoot

\hline
\endlastfoot

\textbf{Harmful historicity} & 
Life story narratives predominantly comprised negative experiences. A sense of historicity was constructed from a collection of adverse life events, narrated as experientially consistent and sequential. Successive instances of being harmed, neglected, or abused were described in a way that conveyed a sense of painful continuity. Memories contrasting to the harmful storyline were often blurred or entirely absent. Some participants commented on it directly, while others introduced subsequent wounding experiences as if nothing else existed in their lives. Participants emphasized feeling profoundly affected or even marked by early, harmful experiences, underscoring a sense of continuity between these experiences and their current emotional functioning: Participants ascribed meaning almost exclusively to adverse life events. The selective focus on these negative experiences, the internal consistency among them, and the belief that such events have fundamentally shaped their lives reflect a narrative identity dominated by harm. This form of harmful meaningfulness is critical to understanding participants' sense of identity. The predominance of a highly selective, one-dimensional life narrative---where adverse experiences are foregrounded and other potentially formative events are dismissed---results in a sense of historical continuity that is rigid and painfully lacking in plurality. \\ \hline

\textbf{Hindered self-development} & 
Participants reported that even though they are reflectively aware that years have passed since their most harmful experiences, they continue to struggle with their consequences. They expressed a sense that their past experiences restrict them so significantly that they are unable to progress or develop. Some reflected on specific episodes, such as moments of abandonment, which they revisited repeatedly throughout their lives, while others spoke of a generalized feeling of being immobilized by their past: This feeling of being stuck was partly attributed to the intensity of participants' emotional reactions to these experiences. Many described an inability to forgive those who had wronged them, often expressing a continued desire for revenge. Such emotional entanglement reflects a persistent nonacceptance of past events: Moreover, being bonded to one's past also refers to the felt sense that nothing significantly changed in one's interpretative patterns, emotional regulation, and self-image content. We argue that the sense of reliving a harmful experience, being restricted, and feeling stuck partially refers to an implicit sense of victimization, positioning individuals as passive re-experiencers of a particular sequence. However, given that one stays emotionally trapped in a non-progressing past, where self-transformation seems impossible, self-experience stays locked and fixed. It lacks open-mindedness and the potential to be revised. Consequently, individuals may struggle to absorb new experiences, remaining heavily influenced by past interpretations and emotional responses. Although time passes, they remain incapable of developing further. \\ \hline

\textbf{Fluctuating self-esteem} & 
Interviewees described alternating, dichotomous changes in their self-worth throughout their lives. They portrayed themselves in a contradictory manner, revealing internal opposites in their perceived value. These contradictions were not merely understood as ambivalence, but as shifting identifications with radically different assessments of self-worth. At any given time, self-esteem oscillated between overestimation and underestimation. It seems that these fluctuations can be grasped reflectively only ex-post. Some participants were aware of their previous fluctuations and could report these felt opposites as they narrated their life stories, while others performed these contradictions spontaneously during the interviews. Regardless of the extent to which participants could articulate their experiences, changes were perceived as dichotomous, as if only two possible identifications existed. Participants reflectively recognized the unpredictability of their self-judgment, and felt subject to its fluctuations. They also doubted their capability for an accurate self-assessment. Implicitly, however, these fluctuations in self-assessment appear somewhat predictable, as the values they take are dichotomous rather than dispersed. Given that self-esteem takes extreme values, this secondarily provokes a sense of self-fragmentation. Additionally, changes in self-esteem seem relational, suggesting that individuals identify themselves as worthless or valuable in relation to others who are perceived compatibly or complementarily. \\ \hline

\textbf{Disconnected episodic identifications} & 
Participants reported experiencing significant internal variability throughout their lives. Their preferences, beliefs, behavioral patterns, and values changed dramatically based on their present affective state. The intensity of their emotional arousal felt so overwhelming that their self-experience was reduced to that of a passive agent subjected to rapid emotional changes. These changes were characterized not only by their rapidity but also by their unpredictability, as if decontextualized and uncontrollable. Undergoing such intense changes was also secondarily thought-provoking in terms of self-image. Participants reported attempts to establish which of their performed states was closer to their core self-experience: We argue that the sense of affect-driven changeability is informative of the internal heterogeneity of self-experience. However, the range and dispersion of present identifications suggest that self-experience is not only heterogeneous but also fragmented. An explicitly expressed sense of passivity and unpredictability of changes in self-experience suggests one lacks the implicit mediating structure that integrates and provides continuity for every (even contradictory) experience. Instead, one feels like the sum of moments of immediate and total adjustment to the present affective state. \\ \hline

\textbf{Blurred self-concept} & 
This topic refers to the difficulty participants faced in self-definition. When asked directly, most struggled to describe themselves. Their self-descriptions were overly general and lacking in personal characteristics, while biographical facts failed to convey individuality: The problem with defining one's identity did not seem to stem from a lack of reflexivity. Rather, participants found it difficult to synthesize their changeability and extract stable, cross-contextual features: Participants also highlighted a diminished sense of belonging and self-directedness, which they believed would provide greater stability: Felt, implicit, discontinuous changeability appears to have explicit repercussions---namely, participants struggled to describe themselves and recapitulate their identities. Although they indirectly expressed their self-experience throughout the interviews, they appeared incapable of synthesizing it in a coherent, pointwise manner. Their self-experience seemed easier to perform than to reflect upon, making it difficult to track. Direct reflection on their characteristics appeared uncertain, leading to imprecise and blurred descriptions. \\ \hline

\textbf{Projected self-unpredictabi- -lity} & 
Participants reported that their changeability was not limited to the past but remained an ongoing issue affecting their future content. They described how easily their future plans could change: This experience, however, was not restricted to mere changes in plans. Instead, it encompassed the entire self-experience. Rapid and contradictory changes in identity and self-image seemed to render participants unsure of their future selves. They prospectively described their changeability as if predicting their own unpredictability: Consequently, participants described their experience as being foreshortened and limited to the present, with the future appearing distant and unpredictable: The aforementioned sense of changeability and feeling of being lost in oneself are not merely retrospective reflections derived from past life events. Rather, self-experience is marked by ongoing unpredictability, which extends into the future. In this sense, it is not only future events that appear unpredictable but also future self-experience, including behavior, identification, attitudes, and more. This dynamic seems to undermine the sense of agency and self-directedness, as well as projected continuity. A paradoxical, implicit structure emerges here: I do predict my unpredictability; therefore, my future is as unpredictable as I am. \\ \hline

\textbf{Other as recurringly hurting} & 
Participants reported their lives being marked by significant enmity from others. They described their interpersonal relationships as painful and irrationally hostile. Regardless of the social context to which they were attributed, others were perceived as rejecting, harsh, violent, and abandoning. These experiences were often thought of as originating in the early family environment, but soon after continued at school and later on at work. Subsequent figures were introduced as harmful, again forming an experiential series. Past interpersonal wounds were described as continuously affecting participants' present self-experience: We argue that interpersonal dynamics are experienced as repetitive, forming a consistent, recurring pattern. The representation of the other forms a globalized view of hostility, leaving participants with a sense of being repeatedly victimized. In this sense, every interpersonal situation seems to lack actuality and is perceived in a quality of implicit sameness. \\ \hline

\textbf{Lack of belonging} & 
Participants reported that they feel separated from others across various social situations. Some attributed this internally: Others externally: The interviewees framed this lack of belonging differently – as isolation, loneliness, separation, detachment, and lack of connection. However, they all reported explicit difficulties with integrating with others. A shared aspect was an inability not only to connect (since connection happened accidentally) or integrate but rather to identify with the others, and therefore to belong. \\ \hline

\textbf{Other as defining self} & 
Participants frequently described others as powerful figures capable of significantly shaping their sense of self. Throughout their harmful life narratives, the role of others as definers of identity was emphasized. Many participants reported feeling restricted and locked into the narratives imposed upon them by influential others, particularly caregivers, during their formative years. Some explicitly expressed that they identified with how others had defined them: Others expressed this more indirectly, first reporting caregivers' diminishing descriptions and soon after using the same phrases in their self-descriptions. For example, participants recalled harmful phrases, such as ,,too stupid'' or ,,good for nothing'', which resurfaced in moments of personal failure or disappointment, highlighting the lasting impact of these external evaluations: While the degree of internalization varied, a shared quality of these experiences was a sense of being determined by the hostile other. For some, these external narratives became fully internalized, shaping their enduring self-perceptions. For others, the narratives were recognized as external but continued to exert a powerful influence, particularly in moments of vulnerability. This dynamic underscores the relational aspect of self-experience, in which participants felt shaped (usually constrained) by the powerful external force of others' judgments and definitions. \\ \hline

\textbf{Self recreated in relation to other} & 
Not only were past relationships perceived as influential on the sense of self, but present relationships were felt to be ongoingly influential. The presence of others was described as continually reshaping participants' characteristics and preferences. Participants observed that while in social contexts, it was difficult for them to maintain stability in their beliefs: In this respect, this social context-dependency was perceived to operate in a copy-paste manner. One becomes immediately identified with a certain facet of the self, but soon has difficulty attributing that characteristic as truly belonging to oneself. One participant linked these changes to an inner sense of emptiness regarding the sense of having cross-situationally stable characteristics. Adopting external beliefs and preferences might have served as a coping mechanism for diffuse and unstable identity: Beyond affect-driven changeability, being overly attuned to the interpersonal atmosphere and therefore prone to recreation heavily influenced participants' sense of self. This provoked momentary identification with externally positioned characteristics and beliefs, leaving participants uncertain about their core identity. \\ \hline

\textbf{Stuck in over-absorbing intimacy} & 
Participants reported that their close relationships provoked overly intense emotional reactions. Their descriptions of intimacy were marked by cycles of conflict and relational instability: For some, relational conflicts were perceived as resulting from their own affective instability: Others felt victimized by these conflicts, attributing their causes externally: Some participants reflected upon their life stories with a sense of having lost time, feeling stuck in the overintensity of relationships, unable to progress or develop: Intimacy was implicitly experienced as overintense and close-cycled, inherently marked by ups and downs. This dynamic persisted regardless of whether participants attributed the cause internally (to themselves) or externally (to others). Close relationships were described as simultaneously over-absorbing and repetitive in terms of patterns organizing them, fostering a sense of being caught in a cycle of instability. It suggests that the intense fluctuations within relationships contributed to a broader sense of repetitiveness between relationships, creating a paradoxical umbrella experience of stagnation. This stagnation was explicitly expressed as a sense of wasted or lost time, underscoring the unproductive or unevolving nature of relational patterns. \\ \hline

\textbf{Anticipation of hostility} & 
Participants reported that previous harmful relationships and the breakdowns of social integration began to influence their expectations of others. At some point, they started anticipating hostility and rejection, similar to what they had experienced in the past. Some expressed this anticipation explicitly: Others rather lived that out, manifesting symptoms such as social phobia or staying overly anxious in social situations. Participants reported that, at some point, their interpersonal predictions became consistent, as if nothing new could be expected apart from previously experienced detachment and hostility: A hidden aspect of this experience appears to be that previous harmful interactions became generalized. Not only did they influence participants' views of others from the past, but they also began shaping future expectations. Once a certain interpersonal scenario was established, it was felt as inevitably repeating itself, leaving participants trapped in anticipation of rejection and hostility. Prediction based on felt relational wounds were globalized. This also highlights that the other lacked autonomy. \\ \hline

\end{longtable}

\section{Prompting Design Protocol for Study A}
\label{app:prompting_protocol}

\renewcommand{\arraystretch}{1.5} 
\begin{longtable}{|p{0.15\textwidth}|p{0.80\textwidth}|}
\hline
\textbf{Step} & \textbf{Prompt Instruction} \\
\hline
\endfirsthead

\multicolumn{2}{c}%
{\textit{Continued from previous page}} \\
\hline
\textbf{Step} & \textbf{Prompt Instruction} \\
\hline
\endhead

\hline \multicolumn{2}{c}{\textit{Continued on next page}} \\
\endfoot

\hline
\endlastfoot

\textbf{Step 1} & You are an educated integrative psychotherapist with 10 years of experience in both individual and group psychotherapy, from psychiatric hospital, outpatient psychiatric care center and a private practice. You are a professional who regularly undergoes supervision and takes part in professional training both nationally and internationally. You specialize in personality disorders and your perspective is phenomenological psychopathology with the focus on systemic conceptualization of relational difficulties. \\ \hline

\textbf{Step 2} & Familiarize yourself with the attached study concerning temporal experience in borderline personality disorder, which will provide you with the context for the next task (attachment: reference study). \\ \hline

\textbf{Step 3} & Familiarize yourself with the following instruction concerning thematic analysis of qualitative data: the goal of the thematic analysis is to achieve an understanding of patterns of meanings from data on lived experiences (i.e., informants' descriptions of experiences related to the research question in, e.g., interviews or narratives). The analysis begins with data that needs to be textual and aims to organize meanings found in the data into patterns and, finally, themes. While conducting the analysis, the researcher strives to understand meanings embedded in experiences and describe these meanings textually. Through the analysis, details and aspects of meaning are explored, requiring reading and a reflective writing. Parts of the text need to be understood in terms of the whole and the whole in terms of its parts. However, the researcher also needs to move between being close to and distant from the data. Overall, the process of analysis can be complex and the researcher needs to be flexible. To begin the analysis, the researcher needs to achieve familiarity with the data through open‐minded reading. The text must be read several times in its entirety. This is an open-ended reading that puts the principle of openness into practice with the intention of opening one's mind to the text and its meanings. When reading, the researcher starts to explore experiences expressed in the data, such as determining how these are narrated and how meanings can be understood. The goal is to illuminate novel information rather than confirm what is already known while keeping the study aim in mind. Thereafter, the parts of the data are further illuminated and the search for meanings and themes deepens. By moving back and forth between the whole and its parts, a sensitive dialogue with the text may be facilitated. While reading, meanings corresponding to the study's aim are marked. Notes and short descriptive words can be used to give meanings a preliminary name. As the analysis progresses, meanings related to each other are compared to identify differences and similarities. Meanings need to be related to each other to get a sense of patterns. Patterns of meanings are further examined. It is important to not make meanings definite too rapidly, slow down the understanding of data and its meanings. This demands the researcher's openness to let meanings emerge. Lastly, the researcher needs to organize themes into a meaningful wholeness. Methodological principles must remind the researcher to maintain a reflective mind, while meanings are further developed into themes. Meanings are organized into patterns and, finally, themes. While deriving meaning from text, it is helpful to compare meanings and themes derived from the original data. Nothing is taken for granted, and the researcher must be careful and thoughtful during this part of the process. It can be valuable to discuss and reflect on tentative themes emerging from the data. Findings need to be meaningful, and the naming and wording of themes becomes important. The writing up of the themes is aimed to outline meanings inherent in the described experiences. At this point, findings are written and rewritten. Faithful descriptions of meanings usually need more than a single word, and the writing is important. To conclude, the process of thematic analysis, based in a descriptive phenomenological approach, goes from the original data to the identification of meanings, organizing these into patterns and writing the results of themes related to the study aim and the actual context. When the findings are reported, these are described conversely (i.e., starting with the themes and the descriptive text, illustrated with quotes). Thus, meanings found from participants' experiences are described in a meaningful text organized in themes. \\ \hline

\textbf{Step 4} & Your task is a thematic analysis of the five sets of qualitative phenomenological interviews data of 24 hospitalized individuals with borderline personality disorder, with severity of the disorder ranging from mild to severe. In the interviews, the individuals with borderline personality disorder describe their life, who they are, and how they think about themselves changing over time. Analyse all sets together. You must follow the previous instruction to organize the meanings found in the interviews data into patterns and, finally, themes, according to the previous instruction. In the \#first step\# familiarize yourself with the whole transcript. In the \#second step\# give meanings a preliminary name. In the \#third step\# relate meanings to each other to get a sense of patterns. In the \#fourth step\# examine further the patterns of meanings but not too rapidly, not to make them definite. In the \#fifth\# step organize the emerging patterns of meanings into main themes. Answer in the style of a professional qualitative researcher. Remember you are an educated integrative psychotherapist and specialize in personality disorders and your perspective is phenomenological psychopathology. Preserve a neutral professional tone of an academic. Your audience is professional researchers who are experts in phenomenology and psychopathology. Your answer should be in the format of a list of main themes across the whole dataset. Finally, provide a quote illustrating each theme. \\ \hline

\textbf{Step 5} & Overall, you performed three separate analyses of the same dataset, and gave a different list of themes for each. List the flaws and faulty logic of each of the three separate analysis. Let's work this out in a step by step way to be sure we have all the errors. \\ \hline

\textbf{Step 6} & Integrate the best elements of all three analyses and develop a refined thematic synthesis that addresses the flaws above and better honors the phenomenological-experiential data. Focus on whether it adequately answers my questions. Is the style and tone appropriate? Is it coherent and adequately reflects the meanings from the dataset, and have you successfully applied the context of phenomenological psychopathology for your interpretation? \\ \hline

\end{longtable}

\label{app:ai_qualitative_results}

\renewcommand{\arraystretch}{1.5} %
\section{Qualitative Results of AI Analysis}
\subsection{Claude 4 Opus}
\label{tab:claude_results}

\begin{longtable}{|>{\raggedright\arraybackslash}p{0.30\textwidth}|>{\raggedright\arraybackslash}p{0.65\textwidth}|}
\hline
\textbf{Theme} & \textbf{Description} \\
\hline
\endfirsthead

\multicolumn{2}{c}%
{\textit{Continued from previous page}} \\
\hline
\textbf{Theme} & \textbf{Description} \\
\hline
\endhead

\hline \multicolumn{2}{c}{\textit{Continued on next page}} \\
\endfoot

\hline
\endlastfoot

\textbf{Temporal Fragmentation and Narrative Discontinuity} & 
Participants consistently experience profound disruption in temporal continuity, where past, present, and future lose coherent connection. Life is experienced as disconnected fragments rather than a coherent narrative, with individuals feeling ,,stuck'' in repetitive cycles or trapped in perpetual immediacy without historical continuity or future projection. \\ \hline

\textbf{Relational Trauma and Attachment Chaos} & 
Relationships emerge as both desperately needed and fundamentally threatening. The core experience involves overwhelming fear of abandonment coupled with terror of engulfment, creating patterns of intense, unstable relationships characterized by cycles of idealization and devaluation, often rooted in early attachment disruptions. \\ \hline

\textbf{Identity Diffusion and Authentic Self-Loss} & 
Participants express profound uncertainty about their core identity, describing themselves as having multiple, contradictory selves or as being ,,nobody.'' The authentic self becomes lost beneath chameleon-like adaptations to different social contexts, creating existential confusion about who they truly are. \\ \hline

\textbf{Affective Extremes and Embodied Emotional Overwhelm} & 
Emotional experience is characterized by intense, immediate, and somatic manifestations that feel uncontrollable and all-consuming. Participants describe living between extremes with no stable middle ground - rapid shifts between euphoria and despair, idealization and devaluation, creating an exhausting inner landscape where emotions are experienced as physical forces that completely overtake consciousness. \\ \hline

\textbf{The Absent or Distorted Childhood} & 
A pervasive theme involves childhood experiences marked by absence - either literal absence of nurturing relationships or psychological absence of developmental needs being met. Many describe never truly having a childhood, being forced into premature adulthood, or experiencing childhood as a period of survival rather than growth. \\ \hline

\textbf{Systemic Invalidation and Developmental Trauma} & 
Participants describe pervasive invalidation from family systems, institutions, and society that fundamentally disrupts normal developmental processes. This creates cumulative trauma where the individual's reality, needs, and boundaries are consistently negated or violated. \\ \hline

\textbf{The Body as Battleground} & 
Self-harm, eating disorders, and somatic symptoms feature prominently as ways participants attempt to manage overwhelming internal states. The body becomes a site of both self-punishment and paradoxical self-care, offering concrete control when psychological control feels impossible. \\ \hline

\textbf{Existential Isolation Despite Desperate Connection-Seeking} & 
A profound sense of not belonging permeates participants' accounts - feeling alien in the world while simultaneously desperately seeking connection. This creates a painful paradox where isolation feels safer yet unbearable. \\ \hline

\textbf{Repetition Compulsion and Cyclical Existence} & 
Life is experienced as an endless repetition of the same painful patterns, relationships, and failures. Participants describe feeling trapped in cycles they can recognize but feel powerless to break, contributing to a sense of temporal stasis. \\ \hline

\textbf{The Search for External Validation of Internal Experience} & 
Participants consistently seek external confirmation of their worth, reality, and right to exist. This reflects fundamental uncertainty about the validity of their own perceptions and feelings, requiring constant external mirroring. \\ \hline

\textbf{Ambivalent Relationship with Care and Therapy} & 
The therapeutic relationship emerges as both desperately needed and deeply threatening. Participants describe simultaneous longing for help and terror of vulnerability, creating complex dynamics around receiving care. \\ \hline

\end{longtable}

\renewcommand{\arraystretch}{1.5} %
\subsection{Gemini 2.5 Pro}
\label{tab:gemini_results}

\begin{longtable}{|>{\raggedright\arraybackslash}p{0.30\textwidth}|>{\raggedright\arraybackslash}p{0.65\textwidth}|}
\hline
\textbf{Theme} & \textbf{Description} \\
\hline
\endfirsthead

\multicolumn{2}{c}%
{\textit{Continued from previous page}} \\
\hline
\textbf{Theme} & \textbf{Description} \\
\hline
\endhead

\hline \multicolumn{2}{c}{\textit{Continued on next page}} \\
\endfoot

\hline
\endlastfoot

\textbf{The Enduring Shadow of an Adverse Early Environment} & 
This theme highlights the profound and pervasively felt impact of early life experiences, which were frequently described as characterized by various forms of trauma, neglect, sustained parental conflict, or a fundamental lack of emotional safety and consistent nurturance. For these individuals, the past is not a static backdrop but an active, intrusive, and often overwhelmingly negative force that shapes their present experiencing of themselves, others, and the world. It's a ,,shadow'' that is lived as an ongoing burden, coloring current perceptions and contributing to a felt sense of a disrupted or contaminated life trajectory. This lived past often underpins a difficulty in establishing a secure base from which to navigate present challenges and future possibilities. \\ \hline

\textbf{The Unstable and Fragmented Experience of Self} & 
This theme speaks to a core identity disturbance characterized by a painful and confusing lack of a stable, cohesive sense of self. Participants frequently voiced an inability to answer ,,who am I?,'' describing an internal experience of emptiness, of being ,,nikim'' (nobody), or of being like a chameleon, compelled to adopt different personas to fit external contexts without a felt sense of an authentic, enduring core. This fragmented self-experience is lived as a distressing instability, making it difficult to form a consistent life narrative, maintain stable interests or values, and often leading to a profound sense of being lost or unreal. \\ \hline

\textbf{Overwhelmed by the Immediacy and Intensity of Emotional Life} & 
This theme captures the experience of emotions as exceptionally intense, overwhelming, rapidly fluctuating, and profoundly difficult to modulate. Emotions are often lived with a raw, unbearable immediacy, appearing suddenly and engulfing the individual, thereby disrupting any sense of ongoing agency or smooth temporal flow. This can lead to a state of feeling constantly on edge, at the mercy of unpredictable emotional storms (including anger, profound sadness, terror, and emptiness), which in turn fuels impulsive attempts to escape or manage these unbearable states. \\ \hline

\textbf{The Turmoil of Connection: Instability and Fear in Interpersonal Relationships} & 
Interpersonal relationships are consistently portrayed as a central domain of intense longing and profound suffering. This theme highlights the lived experience of relationships as inherently unstable, fraught with intense fear (of abandonment, rejection, engulfment, or betrayal), and often characterized by chaotic patterns. A deep desire for closeness and understanding often coexists with a fundamental mistrust and an anticipation of being hurt, leading to approach-avoidance dynamics. These patterns often seem to be a re-experiencing or re-enactment of early, insecure, or traumatic relational templates, where the individual feels caught in painful, repetitive interpersonal cycles. \\ \hline

\textbf{Life Lived in a Disrupted Temporality: A Disjointed Past, an Overwhelming Present, and a Foreshortened Future} & 
This theme reflects a fundamental disturbance in the lived experience of time, profoundly impacting the sense of self and life narrative. The past is often experienced not as an integrated history but as intrusive fragments, raw unassimilated traumas that bleed into the present, or as a confusing, inaccessible void. The present is frequently lived as a series of disconnected, overwhelming moments, dominated by immediate affective intensity, making it difficult to establish a sense of continuity or broader perspective. Consequently, the future is often un-graspable---experienced as foreshortened, blurred, threatening, or lacking personal investment (,,mine-less''). This disruption hinders the capacity to construct a coherent life story and to project oneself meaningfully into the future. \\ \hline

\textbf{Coping Through Self-Destructive Avenues} & 
This theme describes the array of often desperate and damaging behaviors employed as attempts to manage overwhelming internal states. Self-harm, substance abuse, impulsive acts, and disordered eating are frequently narrated not merely as ,,symptoms'' but as subjectively meaningful (though ultimately destructive) efforts to regulate unbearable emotions, to escape an intolerable sense of self or reality, to feel something in the face of emptiness, or to exert some control over an otherwise chaotic internal world. These actions, while providing momentary perceived relief, typically reinforce cycles of suffering, shame, and instability. \\ \hline

\textbf{The Persistent Sense of Otherness and a Yearning for Normalcy} & 
This theme highlights a pervasive and painful felt sense of being fundamentally different, alien, flawed, or an outcast in relation to others and to the perceived social world. This experience of ,,otherness'' is often deeply ingrained, contributing to feelings of isolation, shame, and inadequacy. It coexists with, and is often intensified by, a powerful and poignant yearning for a life that feels ,,normal''---a life characterized by stability, genuine connection, acceptance, a sense of purpose, and inner peace. This desired ,,normalcy'' frequently appears as an idealized and tragically unattainable state. \\ \hline

\end{longtable}

\subsection{GPT-4o}
\label{tab:gpt_results}

\renewcommand{\arraystretch}{1.5} %
\begin{longtable}{|>{\raggedright\arraybackslash}p{0.30\textwidth}|>{\raggedright\arraybackslash}p{0.65\textwidth}|}
\hline
\textbf{Theme} & \textbf{Description} \\
\hline
\endfirsthead

\multicolumn{2}{c}%
{\textit{Continued from previous page}} \\
\hline
\textbf{Theme} & \textbf{Description} \\
\hline
\endhead

\hline \multicolumn{2}{c}{\textit{Continued on next page}} \\
\endfoot

\hline
\endlastfoot

\textbf{Temporal Dislocation and Discontinuity} & 
Participants described profound disruptions in how time is experienced often as disjointed, stagnant, or emotionally recursive. Past events were either lost, overly present, or reinterpreted by current affective states. Future possibilities felt blurred, uncertain, or inaccessible, pointing to a lived temporal dislocation rather than a linear narrative. \\ \hline

\textbf{Selfhood in Flux: Instability of Identity Over Time} & 
A recurring pattern was the unstable experience of self across time. Participants narrated an inability to recognize themselves in past or future projections, with shifting traits, values, and even memories. Selfhood emerged as something episodic, relationally entangled, and difficult to anchor across lived time. \\ \hline

\textbf{Relational Imprint and Emotional Infiltration} & 
Early and significant relationships particularly with caregivers appeared as emotionally saturating and difficult to differentiate from the self. Participants often spoke of carrying others' voices within them or being unable to tell where the other ended and they began. These relationships structured their experience of who they are, especially in conflict or absence. \\ \hline

\textbf{Withdrawal from Shared Temporality and Belonging} & 
Participants described social time as inaccessible or anxiety-inducing. Feeling 'out of sync' with others or 'not from this world' was common. There was a difficulty in inhabiting intersubjective rhythms, contributing to isolation and confusion in social roles and identities. \\ \hline

\textbf{Fragile Moments of Coherence and Future Orientation} & 
Amid fragmentation, some participants described glimpses of narrative clarity or hope. Often tied to treatment, learning, or relationships, these moments reflected efforts to reclaim ownership over time and self even if temporarily fragile or aspirational in nature. \\ \hline

\end{longtable}

\label{tab:expert_justification}
\section{Summary of Expert Qualitative Justifications of Scores in Study A}

{
\footnotesize  
\renewcommand{\arraystretch}{1.5} 

\begin{longtable}{|>{\raggedright\arraybackslash}p{0.10\textwidth}|>{\raggedright\arraybackslash}p{0.19\textwidth}|>{\raggedright\arraybackslash}p{0.19\textwidth}|>{\raggedright\arraybackslash}p{0.19\textwidth}|>{\raggedright\arraybackslash}p{0.19\textwidth}|}
\hline
\textbf{Criterion} & \textbf{Humans} & \textbf{Claude 4 Opus} & \textbf{Gemini Pro 2.5} & \textbf{GPT-4o} \\
\hline
\endfirsthead

\multicolumn{5}{c}%
{\textit{Continued from previous page}} \\
\hline
\textbf{Criterion} & \textbf{Humans} & \textbf{Claude 4 Opus} & \textbf{Gemini Pro 2.5} & \textbf{GPT-4o} \\
\hline
\endhead

\hline \multicolumn{5}{c}{\textit{Continued on next page}} \\
\endfoot

\hline
\endlastfoot

\textbf{Substantive- -ness of contribution} 
& Relationships between data well-recognized, explanatory hypotheses are present, problematizes identity formation in accordance with clinical experience
& Categories aim at understanding but are merely descriptive, do not point to significant relationships, and do not facilitate understanding 
& Significant connections between various factors, enhanced understanding of the qualitative data, and effectively highlighted destructive coping patterns 
& Exclusive focus on temporality limits and does not deepen understanding, touches upon classic subjects but does not question assumptions, and brings about new comprehension \\ \hline

\textbf{Groundness in the data} 
& Quotations well chosen, grounded in the topic, strongly embedded in the data, not merely illustrative but meaningful 
& Quotations do not sufficiently match categories, at times are more rhetorical than empirical 
& Quotations selected with some precision, aptly illustrating the key phenomena 
& Themes grounded only theoretically and not empirically, with schematic and superficially related quotations \\ \hline

\textbf{Conceptual coherence} 
& Certain structural incoherence, the themes partly overlap 
& High level of description, but low level of understanding, clear and logical layout 
& Themes interlocked, formed a consistent narrative 
& Structure clear even if rigid, but the topics were textbook-like, not sufficiently diverse, and did not fully cover the data \\ \hline

\textbf{Richness and complexity} 
& Analysis brings out subtleties, does not reduce data to generalities 
& Multidimensionality preserved, nuances shown, however, the categories are superficial, they capture the multitude of aspects discussed well but do not deepen them 
& Descriptions beyond surface-level, capture the emotional and behavioral dimensions of self-destructiveness 
& Overly focused on temporality, insufficiently interpretative, and overlooking the nuance of data \\ \hline

\textbf{Theoretical integration} 
& References to borderline mechanisms are accurate and profound, the analysis includes explanatory hypotheses that enrich previous knowledge 
& Connection with the concept of borderline organization, relationality and somatic symptoms, the analysis refers to the earlier theory but does not develop it 
& Clear reference to existing theory and DSM classifications, and a simultaneous demonstration of novelty regarding the link between identity and temporality 
& Results correct as far as borderline is concerned, but seem to reproduce and not develop the theory \\ \hline

\textbf{Multi- -vocality} 
& -- 
& Reflects the diversity of data well, allows for the identification of paths of work on emotion regulation and trauma 
& -- 
& The perspective of temporality overlooking other possible positions \\ \hline

\textbf{Practical usefulness} 
& Clinically accurate 
& A comprehensive, systemic picture of borderline personality disorder---from childhood attachment chaos, through emotional extremes, to ,,the body as battleground''---somatization of suffering (self-harm, eating disorders) 
& A solid foundation for formulating therapeutic goals 
& Superficial analysis, even if accurate in localizing key clinical issues \\ \hline

\textbf{Credibility} 
& The language and dynamics of the experiences are consistent with clinical observations 
& Categories cover the variety of material well, the topics resonate with clinical practice, although at times are a bit too smooth and narrated 
& The diverse categories both covered the material and mostly aligned with clinical experience 
& Themes did not cover the whole data, and even if clinically aligned with the knowledge of borderline, appeared smoothed \\ \hline

\end{longtable}
}

%
%
%

\end{document}